\documentclass[runningheads]{llncs}
\usepackage{amssymb} 
\usepackage{amsmath}
\usepackage{hyperref}
\usepackage{graphicx}
\usepackage[ruled,vlined]{algorithm2e}
\usepackage[utf8]{inputenc}
\begin{document}
\title{FedMRL: Data Heterogeneity Aware Federated Multi-agent Deep Reinforcement Learning for Medical Imaging}

\titlerunning{FedMRL}

\author{Pranab Sahoo\inst{1}\orcidID{0000-0002-0784-1908} 
\and Ashutosh Tripathi\inst{2} 
\and Sriparna Saha\inst{1} 
\and Samrat Mondal\inst{1}} 

\authorrunning{Sahoo et al.}
 \institute{Department of Computer Science and Engineering\\ Indian Institute of Technology Patna, India \and
Rajiv Gandhi Institute of Petroleum Technology, India \\
\email{\{pranab\_2021cs25,sriparna,samrat\}@iitp.ac.in}}

%
%
\maketitle              
\begin{abstract}

Despite recent advancements in federated learning (FL) for medical image diagnosis, addressing data heterogeneity among clients remains a significant challenge for practical implementation. A primary hurdle in FL arises from the non-IID nature of data samples across clients, which typically results in a decline in the performance of the aggregated global model. In this study, we introduce FedMRL, a novel federated multi-agent deep reinforcement learning framework designed to address data heterogeneity. FedMRL incorporates a novel loss function to facilitate fairness among clients, preventing bias in the final global model. Additionally, it employs a multi-agent reinforcement learning (MARL) approach to calculate the proximal term $(\mu)$ for the personalized local objective function, ensuring convergence to the global optimum. Furthermore, FedMRL integrates an adaptive weight adjustment method using a Self-organizing map (SOM) on the server side to counteract distribution shifts among clients' local data distributions. We assess our approach using two publicly available real-world medical datasets, and the results demonstrate that FedMRL significantly outperforms state-of-the-art techniques, showing its efficacy in addressing data heterogeneity in federated learning. The code can be found here~{\url{https://github.com/Pranabiitp/FedMRL}}.

\end{abstract}
\section{Introduction}


Deep learning (DL) algorithms have demonstrated significant achievements in medical image analysis tasks~\cite{rajpurkar2017chexnet},~\cite{sahoo2024multi},~\cite{sahoo2022vision},~\cite{sahoo2022computer}. However, creating effective DL-based models typically requires gathering training data from various medical centers, such as hospitals and clinics, into a centralized server. Obtaining patient data from multiple centers presents challenges due to privacy concerns, legal restrictions on data sharing, and the logistical difficulty of transferring large data volumes~\cite{sahoo2023federated}. Researchers have increasingly employed Federated Learning (FL) as a solution, enabling medical image classification with decentralized data from multiple sources while preserving privacy~\cite{yue2023specificity},~\cite{feki2021federated}. Unlike models trained independently at individual sites, FL can leverage a more diverse and extensive dataset, resulting in improved performance and increased generalizability~\cite{mcmahan2017communication},~\cite{lu2022federated}. The efficacy of federated training encounters challenges due to data heterogeneity within local hospital datasets, resulting in performance degradation in real-world healthcare applications. This heterogeneity manifests in various forms: some hospitals may possess more data from patients at early stages, while others primarily collect data from patients with severe conditions, leading to label distribution skew. Additionally, variations in data quantity among hospitals, with larger institutions having more patient data compared to community clinics, contribute to quantity skew. Moreover, differences in imaging acquisition protocols and patient populations further exacerbate feature distribution skew~\cite{yan2023label}. FedAvg, a foundational FL algorithm, while successful in many scenarios, exhibits diminished efficacy in heterogeneous data settings~\cite{mcmahan2017communication}. To address this challenge, FedProx~\cite{li2020federated} introduced a proximal term $(\mu)$ into the conventional optimization objective to penalize large updates in the model parameters. However, selecting an optimal value for the proximal term $\mu$ in FedProx presents a challenge, as traditional methods like trial and error or heuristics may not effectively adapt to heterogeneous data distributions. 

Distribution shifts within each hospital's private dataset often result in scenarios where the global model performs better for certain hospitals but neglects others. The study~\cite{li2019fair} introduced q-fairness optimization problems in FL, where the parameter q guides the loss function to desirable outcomes. Huang et al.~\cite{Huang_2020} focused on fairness and robustness by dynamically selecting local centers for training, but this approach may not be suitable for the medical domain due to limited participant hospitals. Lyu et al.~\cite{lyu2020collaborative} proposed a collaborative fair FL framework to enforce convergence to different models, addressing fairness differently. Another challenge is that existing methods commonly train the global model by minimizing the average training losses of all local clients~\cite{mcmahan2017communication},~\cite{li2021fedbn},~\cite{wang2020tackling}. However, these approaches lack performance guarantees for individual hospitals as they prioritize average training results, leading to divergent performance across participants~\cite{li2020federated}. This issue is exacerbated in real-world scenarios where data from medical centers differ in size and distribution~\cite{hosseini2023proportionally}. Motivated by the aforementioned challenges, we introduce FedMRL, a novel framework that addresses these issues through three distinct components. Our main contributions are:
\begin{itemize}
   \item The FedMRL framework introduces a novel method for calculating adaptive $\mu$ values by leveraging the QMIX algorithm from Multi-agent Reinforcement Learning (MARL). This approach accounts for client-specific factors such as data distribution, volume, and performance feedback, facilitating dynamic regularization adjustments during FL training.
   
   \item We propose integrating a novel loss function into the local objectives of each client to foster fairness among them. This aims to minimize the disparity between each client's loss and the global loss, thereby reducing bias towards a particular client or group.

   \item Our solution involves a server-side adaptive weight adjustment method using self-organizing maps (SOM). This prioritizes contributions from clients with similar data distributions, determined by cosine similarity between local and global models.
 
  \item Experiments on two real-world medical image classification datasets on severe degree of data heterogeneity $(\alpha= 1)$ show that FedMRL outperforms several state-of-the-art FL methods.
  

 
 \end{itemize}

\section{Problem Statement}
Assuming there are H hospitals, each represented by $h \in {[1,2,\cdots,H]}$, and possessing privately labeled data denoted by $D^h$, the aim is to train a generalized global model over the combined dataset $D= \bigcup_{h=1}^{H} $$D^h$. The global objective function is represented in Eq.~\ref{bb}.
\begin{equation}
\arg\min\limits_{w} L(w) = \sum_{h=1}^{H} \frac{|D^{h}|}{|D|} L_{h}(w)
\label{bb}
\end{equation}
The local objective function $L_h (w)$ in client $h$, which quantifies the local empirical loss over the data distribution ${D^h}$, is represented in Eq.~\ref{cc}.
\begin{equation}
L_{h}(w) = \mathbb{E}_{x \sim D^{h}}[l_h(w;x)]
\label{cc}
\end{equation}
In this context, $l_h$ denotes the loss function utilized by client $h$, and $w$ represents the global model parameters. While the above fixed weighted averaging method offers an unbiased global model estimation in the presence of independent and identically distributed (IID) training samples across clients, non-IID distributions, stemming from device and user heterogeneity, lead to slower convergence and reduced accuracy~\cite{zhao2018federated}. To address this challenge, we propose FedMRL, integrating a novel fairness term into the local objective function, dynamically determining the proximal term for each hospital through a MARL-based approach, and employing a self-organizing map-based aggregation method at the server. The final local objective function is represented as shown in Eq.~\ref{b1}.
\begin{equation}
\arg\min\limits_{w} L(w) = \sum_{h=1}^{H} \frac{|D^{h}|}{|D|} L_{h}(w) + \frac{\mu}{2} \lVert w - w^t \rVert^2 + L_{fair}(w)
\label{b1}
\end{equation}

\section{Proposed Framework}
In this section, we provide a comprehensive overview of our proposed approach FedMRL, which consists of three contributions such as calculating adaptive personalized $\mu$ Value, novel loss function, and server-side adaptive weight Aggregation using SOM. The overall algorithm is represented in Algo.~\ref{algo:fedmrl}. The architecture details are shown in Fig.~\ref{fig:1}.


\begin{figure*}
  \centering
  \includegraphics[width=\linewidth]{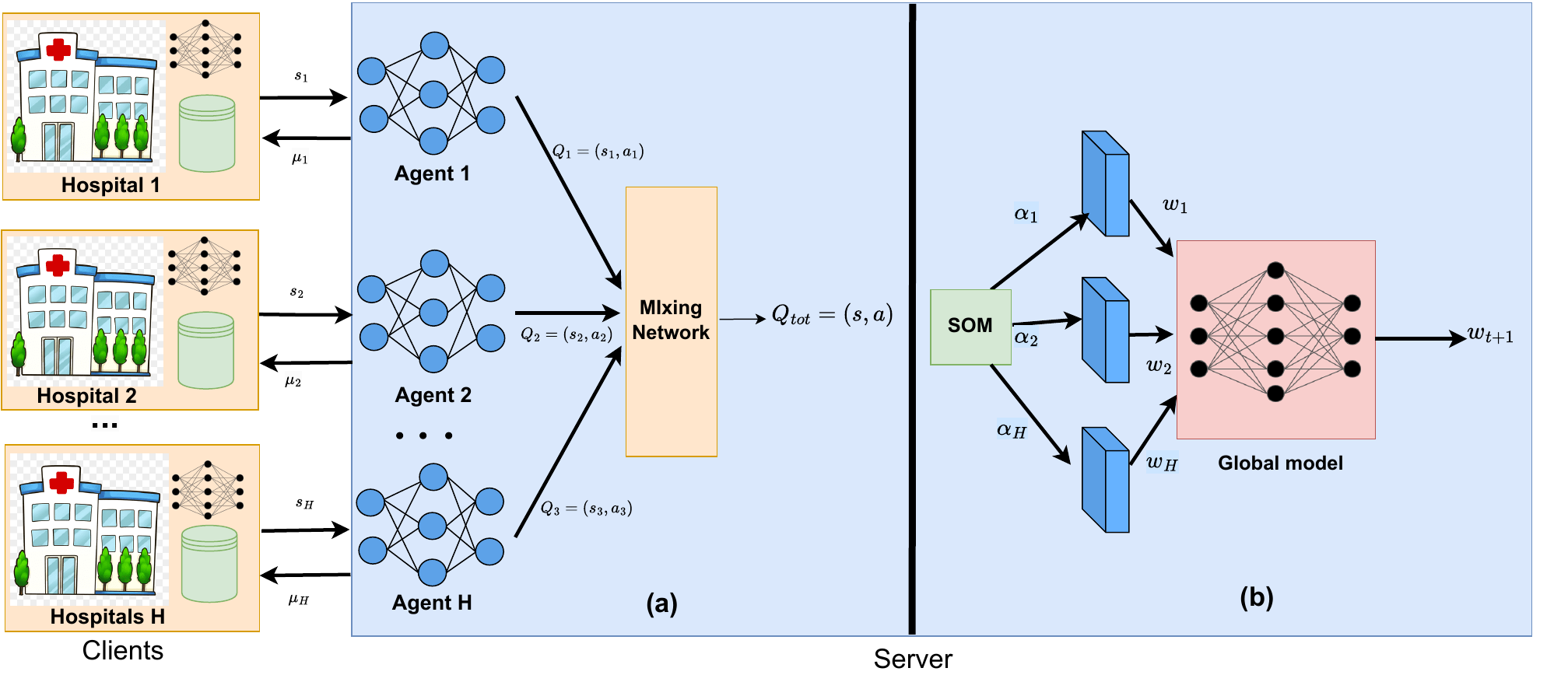}
  \caption{The proposed architecture comprises clients and a server: (a) Clients transmit their model weights to the global server, while agents corresponding to clients retrieve the corresponding state $s_i$ from the global state $s$ and compute the respective $\mu_i$ value, subsequently sharing it with the corresponding hospitals. (b) Global weight aggregation is facilitated using SOM, where $\alpha_i$ denotes the weight adjustment factor, and $w_i$ represents the local model weights utilized to derive the final global model $w_{t+1}$ for subsequent communication rounds.}
  \label{fig:1}
\end{figure*}

\subsection{Adaptive Personalized $\mu$ Value}
\label{apv}

For the dynamic adaptation of the proximal term $\mu$, we frame it as a multi-agent reinforcement learning problem, with each hospital $h$ having an agent on the server side. Each agent $h$ observes its state $s_i$ from the overall environment state $s_t$, selects an action $a_i$ based on the current policy $\pi_t$, and the agents' actions collectively form the joint action $A_i$. The environment transitions to the next state $s_{i+1}$ according to the state transition function $P(s_{i+1}|s_i,a_i)$, iterating until completion or predefined criteria are met. In the proposed work, we integrate QMIX~\cite{rashid2020monotonic}, a prominent Q-learning algorithm for cooperative MARL in the decentralized paradigm that represents an advancement over Value-Decomposition Networks (VDN)~\cite{sunehag2017valuedecomposition}. Essentially, VDN assesses the influence of each agent on the collective reward, assuming that the joint action-value function $Q_{\text{tot}}(s, a)$ can be decomposed into $N$ $Q$-functions for $N$ agents, with each $Q$-function relying solely on local state-action history, represented in Eq.~\ref{ee}.
\begin{equation}
   Q_{\text{tot}}(s, a) = \sum_{j=1}^{N} Q_j(s_j, a_j,\theta_j) 
\label{ee}
\end{equation}
\textbf{State:} The state of the environment at round $t$ is $s_t=[s_{t,1},s_{t,2},\cdots,s_{t,h}]$ represents the data distribution among clients and performance feedback $s_{t,i}$ is defined in Eq.~\ref{ff}.
\begin{equation}
\label{ff}
  \mathbf{s_{t,i}} = (E_c, P_c, \text{$acc_c$}, \text{loss}_c)
\end{equation}
We have represented the entropy of the dataset in the $c^{th}$ client $(E_c)$ as defined in Eq.~\ref{gg} and the ratio of images in the $c^{th}$ client to images in all clients $(P_c)$, are represented in Eq.~\ref{ggg}~\cite{yue2023specificity}. 
\begin{equation}
\label{gg}
E_c = \sum_{m=0}^{M-1} \frac{N_m}{N_c}  \log\left(\frac{N_m}{N_c}\right)
\end{equation}
\begin{equation}
\label{ggg}
  P_c = \frac{N_c}{N} 
\end{equation}
$N_m$, $N_c$, and $N$ represent the number of data in the $c^{th}$ client that belong to the $m^{th}$ class, the total number of data samples in the $c^{th}$ client, and the total data samples in FL sourced from local clients, respectively. Additionally, the $c^{th}$ local model's accuracy and training loss, denoted as $acc_c$ and $loss_c$, respectively,  are dynamic metrics that evolve across communication rounds, serving as performance feedback for the local models. $(E_c)$ and $(P_c)$ depict the features of the local datasets and remain constant when training the network. \\
\textbf{Action:} We consider the proximal term value $\mu$ as our action, a continuous value ranging from $0$ to $1$, and initialize it at a minimal value ($\approx 0.00001$) at the beginning of FL training. Subsequently, following the first communication round, agents determine the proximal term value based on the current policy, resulting in a joint action denoted as $A_t=[a_{t,1}, a_{t,2},\cdots, a_{t,h}]$, where $a_{t, i}$ represents the proximal term value assigned to agent $h_i$. \\
\textbf{Reward:} The observed reward value in round $t$ is denoted as $r_t = e^{{acc_t}-\zeta }- 1$, where $e$ represents the natural constant, $\zeta$ denotes the target accuracy, and $acc_t$ signifies the global model's test accuracy. The QMIX agent maximizes the anticipation of the cumulative discounted reward ($R$) through training, as described in Eq.~\ref{ccc}. $\gamma$ $\in$ (0, 1] is the discount factor.
\begin{equation}
R = \sum_{t=1}^{T} \gamma_{t-1} r_{t} = \sum_{t=1}^{T} \gamma_{t-1} \left( e^{{(acc_t}-\zeta)} - 1 \right)
\label{ccc}
\end{equation}

\subsection{Loss Function}
Our proposed novel loss function in FedMRL effectively mitigates the impact of distribution shifts in hospitals' datasets, ensuring fairness and consistent performance across all participating institutions. By incorporating a fairness term into the local objective function of individual hospitals, FedMRL adjusts model parameters to achieve uniform training loss across all $H$ hospitals, inspired by the Mean Square Error (MSE) loss function commonly used in regression models. The fairness term is formulated as an optimization problem aiming to minimize the sum of squares of differences in loss between each of the $H$ hospitals and the global loss, as presented in Eq.~\ref{gl}.
\begin{equation}
    {L_{fair}} = \sum_{h=1}^{H} \left( F_{h}(w) - F(w) \right)^{2}
\label{gl}
\end{equation}
where, $w$ represents the parameter of the global model, $H$ denotes the number of hospitals, $F_{h}(w)$ signifies the local loss function of individual hospital $h$ based on its local data, and $F(w) = \sum_{h=1}^{H} F_h(w)$ represents the global loss function. The proof of the above-proposed loss function can be found in Section 6 of the
Appendix.

\begin{algorithm}[H]
    \caption{Steps of the proposed FedMRL.}
    \label{algo:fedmrl}
    \SetAlgoLined
    \KwIn{Set of hospitals $H$, initial global model weights $w^0$, target accuracy $\zeta$, number of communication rounds $T$}
    \KwOut{Global model parameter $w_t$}
    \BlankLine
    Initialize $i=0$\;
    \While{$i < T$}{
        Each hospital downloads the weights of the initial global model, performs local SGD training for $1$ epoch by minimizing the local objective function represented in Eq.~\ref{b1}, and uploads the local model weights to the server\;
        Execute QMIX agent as described in Subsection~\ref{apv}\;
        Execute SOM as described in Subsection~\ref{sswa}\;
        Share $w^{t+1}$ back to all the clients for the next communication round\;
        $i \leftarrow i + 1$\;
    }
\end{algorithm}

\subsection{Server side Weight Aggregation}
\label{sswa}
FedAvg uniformly averages client model updates, neglecting individual data distributions, which can impede performance in non-IID scenarios. In contrast, SOM-based weight adjustment considers client model similarity to the global model, significantly impacting clients with more representative data~\cite{kohonen1990self}. This adaptivity effectively addresses non-IID distribution challenges, allowing personalized adjustments based on model-global similarity, thus enhancing performance for clients with unique data distributions. We initialize the SOM grid shape as (5,5), and its weights are randomly initialized. Distances between the SOM weights and each hospital's local weights determine the Best Matching Unit (BMU) on the SOM grid. The influence of local weights on each SOM neuron is calculated based on its distance to the BMU and current sigma value, updating the neuron weights accordingly. Weights for each local model are computed from the SOM weights and similarity metrics, ensuring accurate representation through normalization. Cosine similarity metrics between local and global models, combined with distances, determine weights($\alpha_i)$ for each local model, favoring higher similarity for increased weight. During SOM weight updates, the influence of each local model scales by its similarity metric, with higher similarity models exerting greater impact. Normalized similarity metrics ensure proportional weighting, are responsive to changes in local models over time, and maintain fairness in highly non-IID data settings. The aggregation is performed according to Eq.~\ref{pp}, where $w_{t+1}$ denotes the global model parameters for the $(t+1)^{th}$ communication round, $w_t^h$ signifies the local model weights of the $h^{th}$ hospital, and $\alpha_h$ represents the weight factor for the $h^{th}$ hospital.


\begin{equation}
w_{t+1} = \frac{1}{H} \sum_{h=0}^{H-1} \omega_t^h \cdot \alpha_h
\label{pp}
\end{equation}

\section{Dataset and Experimental Results}
In this section, we present the datasets used for the experiment and the experimental results of the proposed FedMRL.

\subsection{Datasets}

We have chosen two distinct benchmark datasets pertinent to real-world medical contexts to evaluate the effectiveness of our proposed FedMRL framework for highly heterogeneous scenarios. The ISIC 2018 dataset, notable for its contributions to skin cancer detection, provides a diverse array of dermoscopy images captured from various anatomical regions. This dataset encompasses 7,200 images categorized into 7 distinct classes~\cite{codella2019skin}. Additionally, we utilized the Messidor dataset~\cite{decenciere2014feedback}, consisting of 1,560 authentic fundus images tailored for grading diabetic macular edema across five severity levels. Both datasets were partitioned into 80\% for training and 20\% for validation to ensure robust evaluation.

\subsection{Implementation Details}
We construct non-IID data partitions following the methodology outlined in~\cite{mcmahan2023communicationefficient}. Using 80\% of each dataset for training purposes, we organize the data based on their labels and segment each class into 200 shards. Subsequently, clients create local datasets by sampling from these shards according to the probabilities represented in Eq.~\ref{dd}.
\begin{equation}
\text{pr}(x) = \begin{cases}
\eta \in [0, 1], & \text{if } x \in \text{class } j, \\
N(0.5, 1), & \text{otherwise.}
\end{cases}
\label{dd}
\end{equation}
The client samples from a specific class $j$ with a constant probability $\eta$, while samples from other classes follow a Gaussian distribution. Higher $\eta$ values indicate a greater concentration of samples in a specific class, resulting in more heterogeneous datasets. We have used $\eta=1.0$ for the experiment to access the performance of FedMRL. Following~\cite{wang2021deep}, We employ DenseNet121~\cite{huang2017densely} as the backbone. For Fedprox, we selected the proximal term $\mu$ from the set ${0.001,0.1,0.4}$, and for Fednova, we chose the proximal SGD value from the set ${0.001,0.1,0.2}$ to yield the best results.




\begin{table}
\caption{Results comparison with state-of-the-art methods on two datasets.}
\label{table1}  
\centering
\resizebox{0.70\textwidth}{!}{
\begin{tabular}{|c|c|c|c|c|c|c|}
\hline
Dataset &  Method & ACC & AUC  & Pre & Recall & F1\\
\hline
 & Fedavg~\cite{mcmahan2017communication} &72.82 &89.85 &74.69 &72.82&73.74 \\
& Fedprox~\cite{li2020federated} & 72.90& 90.26&74.30&73.21&73.75

 \\
ISIC-2018 & Fednova~\cite{wang2020tackling} & 67.88&86.60&70.51&67.88&
69.17

  \\
& FedBN~\cite{li2021fedbn} &71.71 &90.18&73.46&71.71&72.57

 \\
 & \textbf{FedMRL} & \textbf{73.50} &\textbf{80.01}&\textbf{77.04}&\textbf{71.90}&\textbf{74.38} \\
 \hline
& Fedavg~\cite{mcmahan2017communication} &55.83&86.07&51.96&55.83&53.82

 \\
& Fedprox~\cite{li2020federated} & 57.08&85.99&56.71&57.08&56.89

  \\
Messidor& Fednova~\cite{wang2020tackling} & 50.00&82.21&50.70&50.00&50.34

  \\
& FedBN~\cite{li2021fedbn}
& 55.83&85.07&57.10&55.83&56.45

 \\
& \textbf{FedMRL} &\textbf{57.91} &\textbf{85.82}&\textbf{58.23}&\textbf{57.91}&\textbf{58.06}

 \\
\hline

\end{tabular}
}
\end{table}

\subsection{Comparison with State-of-the-arts}
To assess the efficacy of the proposed FedMRL methods, we compare them against four baseline approaches: FedAvg, FedProx, FedNova, and FedBN. Table~\ref{table1} presents the results of all the models in terms of Accuracy (ACC), Area under the ROC Curve (AUC), Precision (Pre), Recall, and F1-score (F1). The results highlight FedMRL's superior performance over baseline algorithms, demonstrating significant accuracy improvements across both datasets. Specifically, FedMRL outperforms state-of-the-art methods by 0.92\%, 0.81\%, 7.64\%, and 2.43\% for the ISIC-2018 dataset, and by 3.59\%, 1.43\%, 13.65\%, and 3.59\% for the Messidor dataset, compared to FedAvg, FedProx, FedNova, and FedBN, respectively. These improvements benefit from our FedMRL scheme, which explicitly takes advantage of MARL to learn the data heterogeneity in the network and adaptively optimize the local objective of clients. Additionally, our novel loss function promotes fairness among clients, while the SOM-based adaptive weight adjustment method for aggregation enhances convergence to a better global optimum.

\section{Conclusion}
This study addresses the challenge of data heterogeneity in federated learning while ensuring fair contributions from the decentralized participants. Our framework demonstrates robustness to non-IID data distribution across clients and outperforms existing benchmarks in two medical datasets. While effective in mitigating the challenges posed by non-IID data distributions, FedMRL may encounter scalability issues when dealing with a large number of clients. Furthermore, the computational overhead associated with calculating personalized $\mu$ values and performing server-side adaptive weight aggregation using SOM may impose additional computational burdens, particularly in resource-constrained environments. In future work, we envision exploring distributed computing and resource-sharing models to alleviate the computational burden and aim to enhance FedMRL's scalability and efficiency while exploring its applicability in diverse domains. 

\bibliographystyle{splncs04}
\bibliography{sample}

\section*{Appendix}
\label{sec:appendixA}
\section{Proof of Proposed Loss Function}
\label{pplf}
When all $H$ hospitals exhibit identical training loss, the optimal solution for the fairness term emerges. To demonstrate this, we minimize $L_{fair}$. Subsequently, we elaborate on the mathematical representation of the fairness term. Further, we discuss the enhancement in performance resulting from integrating the fairness term into the local training loss, as described by Eq.~\ref{gl}.


\begin{equation}
    {L_{fair}} = \sum_{h=1}^{H} \left( F_{h}(w) - F(w) \right)^{2}
\label{gl}
\end{equation}
\begin{equation}
\begin{split}
     {L_{fair}} = \sum_{h=1}^{H} (F_{h}^{2}(w) + F(w)^{2} - 2F_{h}(w)F(w)) \\
    = \sum_{h=1}^{H} (F_{h}^{2}(w) - 2F_{h}(w)F(w)) + HF(w)^{2} 
    \end{split}
\label{eq:first_expansion}    
\end{equation}
To minimize the loss function, we computed the derivative with respect to $F_j(w)$ for a given hospital $j \in {1,2, \cdots,H}$. Setting it to zero, as represented in Eq.~\ref{aa}, enables us to identify the minima.
\begin{equation}
\begin{split}
    \frac{dL_{fair}}{dF_j(w)} = 2F_j(w) = 0 \\
    F_j(w)=0\\
\end{split}
\label{aa}
\end{equation}
The minimization of $L_{fair}$ implies that $F_h(w)=0$ holds true for all hospitals, which is equivalent to the condition $F_h(w)=F(w)$ for all $h \in {1,2,3,...,H}$. This indicates that all $M$ hospitals will converge to the same local loss, ensuring fairness.
We have shown our proposed loss function by plotting hospital 1’s loss $(F_1)$ while fixing the total loss. The plot (refer to Fig.~\ref{fig:1}) confirms the quadratic penalty effect - the loss grows with the divergence between F1 and the fair 0.5 value.
Importantly, the plot also shows that the global minimum is precisely when $F_1 = F_2$. The equal losses satisfy the condition for fairness we derived mathematically in the proof.
That clearly means all clients will converge to the same local loss, satisfying fairness across all $H$ hospitals.

\begin{figure*}
  \centering
  \includegraphics[width=0.7\linewidth]{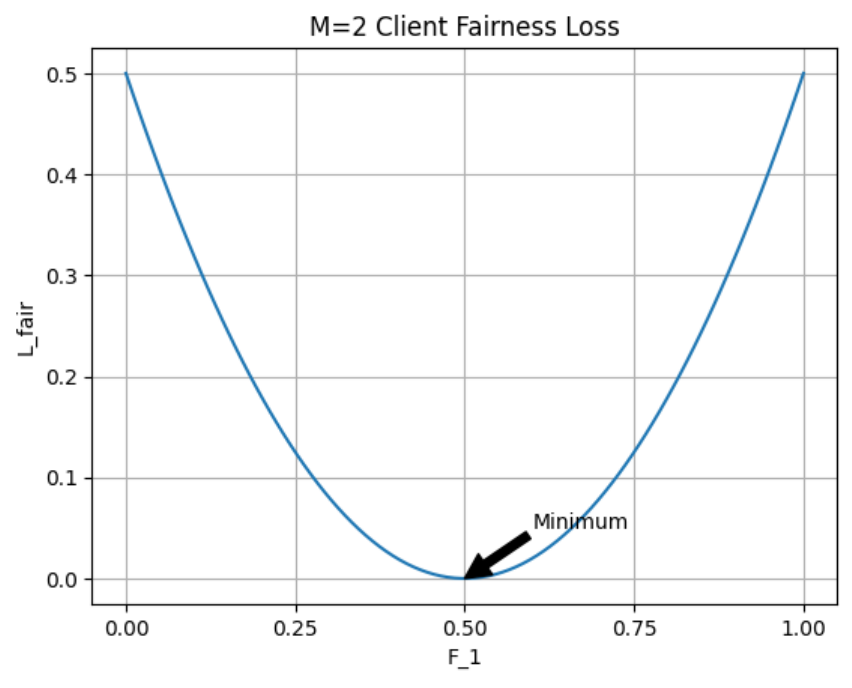}
  \caption{The plot of the proposed loss function considering the 2 clients. It is clear from the graph that the minimum of the loss function occurs when both clients have the same loss value, i.e., tending to zero.}
  \label{fig:1}
\end{figure*}

%




\end{document}